%% file: main.tex
\documentclass[10pt,twocolumn,letterpaper]{article}

\usepackage{cvpr}              

\input{preamble}

\definecolor{cvprblue}{rgb}{0.21,0.49,0.74}
\usepackage[pagebackref,breaklinks,colorlinks,citecolor=cvprblue]{hyperref}

\def\paperID{5729} 
\def\confName{CVPR}
\def\confYear{2024}

\title{Neural Refinement for Absolute Pose Regression with Feature Synthesis}

\author{Shuai Chen$^1$ \qquad Yash Bhalgat$^2$ \qquad Xinghui Li$^1$ \qquad Jiawang Bian$^1$ \\
Kejie Li$^1$ \qquad Zirui Wang$^1$ \qquad Victor Adrian Prisacariu$^1$ \\
$^1$Active Vision Lab, University of Oxford\\
$^2$Visual Geometry Group, University of Oxford
}

\begin{document}
\maketitle

\input{sec/00_abs}

\input{sec/01_intro}

\input{sec/02_related_work}
\input{sec/03_method}

\input{sec/04_exp}
\input{sec/05_conclusion}


\input{supp/supp_00}

{
    \small
    \bibliographystyle{ieeenat_fullname}
    \bibliography{main}
}


\end{document}

%% file: preamble.tex
\usepackage[dvipsnames]{xcolor}
\usepackage{epsfig}
\usepackage{amsmath}
\usepackage{amssymb}
\usepackage{booktabs}
\usepackage{comment}
\usepackage{color} 
\usepackage{epsfig}
\usepackage{textcomp}
\usepackage{gensymb}
\usepackage{makecell}
\usepackage{enumitem}
\usepackage{subcaption}
\usepackage{multirow}

\newcommand\boldred[1]{\textcolor{black}{\textbf{#1}}}

%% file: sec/00_abs.tex
\begin{abstract}
    Absolute Pose Regression (APR) methods use deep neural networks to directly regress camera poses from RGB images. However, the predominant APR architectures only rely on 2D operations during inference, resulting in limited accuracy of pose estimation due to the lack of 3D geometry constraints or priors. In this work, we propose a test-time refinement pipeline that leverages implicit geometric constraints using a robust feature field to enhance the ability of APR methods to use 3D information during inference. We also introduce a novel Neural Feature Synthesizer (NeFeS) model, which encodes 3D geometric features during training and directly renders dense novel view features at test time to refine APR methods. To enhance the robustness of our model, we introduce a feature fusion module and a progressive training strategy. Our proposed method achieves state-of-the-art single-image APR accuracy on indoor and outdoor datasets. Code will be released at \url{https://github.com/ActiveVisionLab/NeFeS}. 


\end{abstract}

%% file: sec/01_intro.tex
\section{Introduction}

Camera relocalization is a crucial task that allows machines to understand their position and orientation in 3D space. It is an essential prerequisite for applications such as augmented reality, robotics, and autonomous driving, where the accuracy and efficiency of pose estimation are important.
Recently, Absolute Pose Regression (APR) methods \cite{Kendall15,Kendall16,Kendall17} have been shown to be effective in directly estimating camera pose from RGB images using convolutional neural networks. The simplicity of APR's architecture offers several potential advantages over classical geometry-based methods \cite{Sattler17, sarlin21pixloc, brachmann2020dsacstar}, involving end-to-end training, cheap computation cost, and low memory demand. 

Latest advances in APR, particularly the use of novel view synthesis (NVS) \cite{Mildenhall20, martinbrualla2020nerfw, Shavit22PAE, chen21, chen2022dfnet, Moreau21} to generate new images from random viewpoints as data augmentation during training, have significantly improved the pose regression performance.
Despite this, state-of-the-art (SOTA) APRs still have the following limitations: 
(i) 
They predict the pose of a query image by passing it through a CNN, which typically disregards geometry at inference time.
This causes APR networks to struggle to generalize to viewpoints that the training data fails to cover \cite{Sattler19};
(ii) The unlabeled data, often sampled from the validation/testing set, used for finetuning the APR network~\cite{Brahmbhatt18, chen21, chen2022dfnet} may not be universally available in real-life circumstances, and this semi-supervised finetuning is also time-consuming.
\input{fig_tex/figure0_teaser}

To address these limitations, we propose a novel test-time refinement pipeline for APR methods. Unlike prior works that
explore extended Kalman filters \cite{moreau2022coordinet}, pose graph optimization \cite{Brahmbhatt18}, or pose auto-encoders \cite{Shavit22PAE}, our method integrates an \textit{implicit} representation based geometric refinement into an end-to-end learning framework, where gradients can be backpropagated to the APR network. We test our proposed method across different APR architectures to demonstrate its robustness and effectiveness. 
Furthermore, we propose a Neural Feature Synthesizer (NeFeS) network to encode the 3D geometry of a scene into an MLP. NeFeS render dense features from novel viewpoints for refinement. To ensure the robustness of feature rendering, we introduce a Feature Fusion module into NeFeS that combines the rendered color and features and is trained in a progressive manner.
Our method leverages prior literature on volume rendering to inherently constrain geometric consistency during test time using implicit 3D neural feature fields. As such, our approach occupies a middle ground between APR and methods informed by geometry.

We summarize our main contributions as follows:
\textbf{First}, we propose a test-time refinement pipeline that greatly improves the pose-estimation accuracy of any APR model without using additional unlabeled data and exhibits a new \textit{single-frame} APR SOTA performance on standard benchmarks.
\textbf{Second}, we propose a Neural Feature Synthesizer (NeFeS) network that encodes 3D geometric features. NeFeS refines an initial pose by rendering a dense feature map and making the comparison with the query image feature.
\textbf{Third}, we propose a progressive training strategy and a Feature Fusion module to improve the robustness of the rendering ability of the NeFeS model.

%% file: fig_tex/figure0_teaser.tex
\begin{figure}[t]
    \centering
    \includegraphics[width=\linewidth]{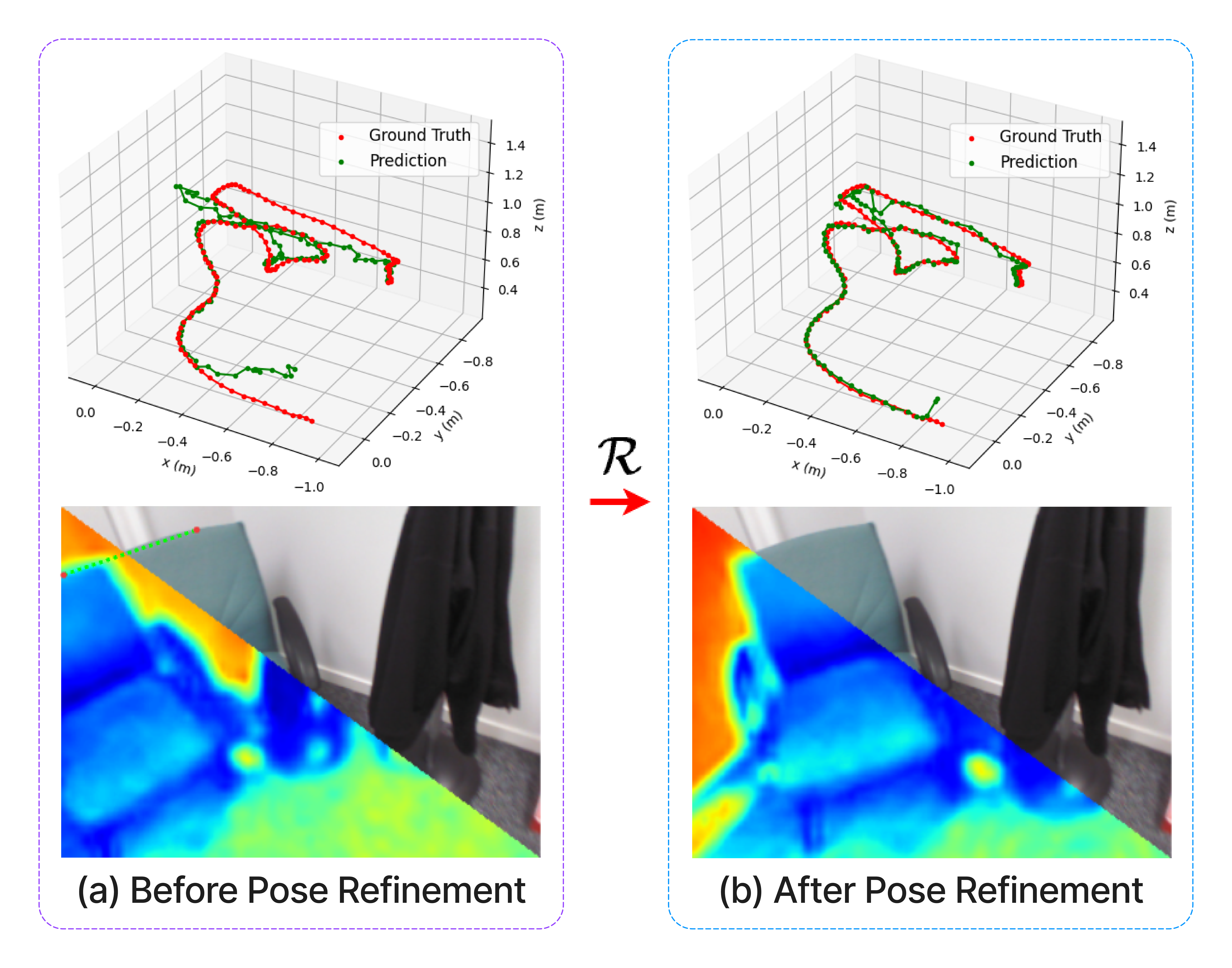}
    \caption{Our pose refinement ($\mathcal{R}$) improves (coarse) pose predictions from other methods using novel feature synthesis to achieve pixel-wise alignment. \textbf{Top left / right:} 3D plots of predicted (\textcolor{Green}{\textbf{green}}) and ground-truth (\textcolor{Red}{\textbf{red}}) camera positions.
    \textbf{Bottom left / right:} alignment between rendered features and query image.}
    \label{fig:teaser1}
\end{figure}

%% file: sec/02_related_work.tex
\section{Related Work} \label{sec:related_work}

\textbf{Absolute Pose Regression (APR).} APR methods have been widely studied due to their simple and lightweight formulation that allows the camera pose to be directly regressed using an end-to-end neural network. PoseNet \cite{Kendall15, Kendall16, Kendall17} introduced the first APR solution using GoogLeNet-backbone, followed by various architectures like the hourglass network \cite{Melekhov17}, attention layers \cite{atloc,Shavit21multiscene,Shavit21}, separated translation and rotation prediction \cite{Wu17,Naseer17}, or LSTM \cite{Walch17}.

To further improve APR accuracy, some works utilize sequential information. These approaches incorporate temporal constraints such as visual odometry \cite{Brahmbhatt18, Valada18, Radwan18}, motion \cite{moreau2022coordinet}, temporal  filtering \cite{clark2017vidloc}, and multi-tasking \cite{Radwan18}. Recent APR methods also benefit from novel view synthesis, where one line of approaches focuses on generating large amounts of extra photo-realistic synthetic data \cite{Purkait18, Moreau21, chen2022dfnet} via randomly sampled virtual camera poses. However, generating high-quality offline synthetic data may take up to several days \cite{Moreau21} for each scene. Other approaches \cite{chen21, chen2022dfnet} use NeRF \cite{Mildenhall20,martinbrualla2020nerfw} as a direct matching module to perform unlabeled finetuning \cite{Brahmbhatt18} using extra images without ground-truth pose annotation. However, finetuning usually takes significant time and assumes that extra unlabeled data can be easily obtained.

While the aforementioned works enhance APR training, we focus on improving generic APR methods during test time. 
Unlike prior works that only exam means for test-time refinement on a single specific APR architecture, such as extended Kalman filters \cite{moreau2022coordinet}, pose graph optimization \cite{Brahmbhatt18}, or pose auto-encoders \cite{Shavit22PAE}, our method exhibits strong flexibility to be adapted to a wide range of APR architectures on both camera positions and orientations, achieving state-of-the-art results without extra unlabeled data.

Notably, classical geometry-based techniques \cite{Sattler12, Sattler17, sarlin21pixloc, sarlin2019HFNet, Lindenberger21, Brachmann17, Brachmann18, brachmann2020dsacstar} that require explicit feature correspondence search \cite{DeTone18, Sarlin20, Taira18, sun2021loftr, Li20, dusmanu2019d2} also employ test-time refinement to improve localization accuracy. For example, \cite{sarlin2019HFNet,sarlin21pixloc, Lindenberger21} build upon image retrievals and pre-computed SfM model to perform standard geometric refinement via neural network-based feature matcher, PnP+RANSAC, or dense featuremetric-alignment. Our method, however, offers end-to-end neural feature refinement via implicit representation, enhancing existing APR models without external storage, pre-computed data, or manual tuning.

\input{fig_tex/figure1_pipeline}
\textbf{Neural Radiance and Feature Fields.} Neural Radiance Fields (NeRF) \cite{Mildenhall20} revolutionized novel view image synthesis and 3D surface reconstruction. NeRF's implicit 3D representation and differentiable volume rendering enable self-supervised optimization from RGB images, avoiding costly 3D annotations. iNeRF \cite{yen2020inerf} showed that NeRF can be inverted for pose optimization. Recent approaches such as BARF \cite{lin21barf} and its counterparts \cite{wang2021nerfmm,Chng22garf,bian22nope} simultaneously train NeRF by treating camera poses as learnable parameters in simple, non-360\degree scenes. Parallel works,  NICE-SLAM \cite{niceslam} and iMAP \cite{imap}, use NeRF for dense geometry and real-time camera tracking. Direct-PN \cite{chen21} uses NeRF as a direct matching module to compute the photometric errors and propagate the error gradients back to the pose regression network. DFNet \cite{chen2022dfnet} extends this method to outdoor scenarios with robust feature extraction. LENS \cite{Moreau21} uses NeRF to generate a synthetic training dataset based on manually tuned scene bounds and parameters.

Recently, NeRF models have been extended to directly predict and render \textit{feature fields} alongside density and appearance fields. Typically, these feature fields are learned by supervision from a 2D feature extractor using volumetric rendering. \cite{neff,Kobayashi22,bhalgat2023contrastive} showed that these 3D feature fields outperform 2D baselines \cite{dino,clip_lseg,cheng2021mask2former} on downstream tasks such as 2D object retrieval or 3D segmentation. CLIP-Fields \cite{clip_fields} established feature fields as scene memory for robot navigation. This work explores distilled neural feature fields for camera relocalization, highlighting their role in test-time pose refinement.

%% file: fig_tex/figure1_pipeline.tex
\begin{figure*}[ht]
    \centering
    \includegraphics[width=.88\linewidth]{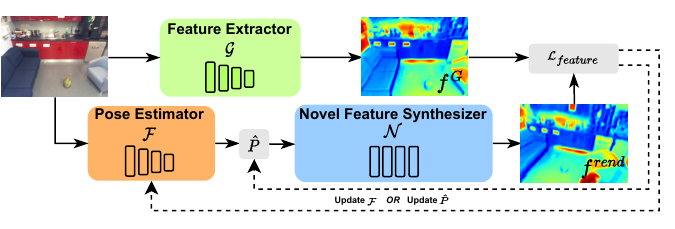}
    \caption{\textbf{Illustration of the pose refinement pipeline.} The query image is processed by a pose estimator $\mathcal{F}$, typically an absolute pose regressor, to obtain a coarse camera pose $\hat{P}$. Our novel feature synthesizer $\mathcal{N}$ renders a dense feature map $f^{rend}$ based on $\hat{P}$. Simultaneously, the feature extractor $\mathcal{G}$ extracts the feature map $f^{G}$ from the query image. We then compute the feature-metric error between $f^{rend}$ and $f^{G}$, denoted as $\mathcal{L}_{feature}$. This error is backpropagated to update either the parameters of $\mathcal{F}$ or the coarse pose $\hat{P}$ directly.}
    \label{fig:pipeline}
    \vspace{-10pt}
\end{figure*}

%% file: sec/03_method.tex
\section{Method} \label{sec:method}
In this section, we present a detailed outline of our approach. \cref{sec:PP_APR} provides a high-level overview of our refinement framework. \cref{sec:NFS} describes the architecture and training details of our proposed NeFeS network along with its two components: \textit{Exposure-adaptive Affine Color Transformation (ACT)} and \textit{Feature Fusion module}. 

\subsection{Refinement Framework for APR} \label{sec:PP_APR}
Given a query image $I$, an absolute pose regression (APR) network $\mathcal{F}$ directly regresses the camera pose $\hat{P}$ of $I$: $\hat{P} = \mathcal{F}(I)$.
The network is typically trained with ground truth image-pose pairs. 
While APR-based methods are much more efficient than geometry-based methods since they require only a single forward pass of the network, the quality of their predictions is often significantly worse than those of geometry-based methods due to the lack of any 3D geometry-based reasoning~\cite{Sattler19}.

In contrast to prior APR research, which attempts to improve APR by adding constraints to the training loss or making architectural changes to the backbone network, we propose an alternative method to refine the results of APR methods by backpropagating a feature-metric error at inference time. 
Our method has three major components (see \cref{fig:pipeline}): (1) a pretrained APR network, denoted as $\mathcal{F}$, which provides an initial pose; (2) a differentiable novel feature synthesizer $\mathcal{N}$ that directly renders dense feature maps given a camera pose; (3) an off-the-shelf feature extractor $\mathcal{G}$ that extracts the dense feature map of the query image. In our implementation, the feature extraction module from \cite{chen2022dfnet} is employed as the feature extractor $\mathcal{G}$.

The refinement procedure is as follows: (i) The query image $I$ is passed through the pretrained APR model $\mathcal{F}$ to predict a coarse camera pose $\hat{P}$.
(ii) The feature synthesizer $\mathcal{N}$ renders a dense feature map $f^{rend}\in \mathbb{R}^{n\times c}$ given the coarse camera pose $\hat{P}$, where $n=h\times w$, and $h$ and $w$ are the spatial dimensions of the feature map\footnote{Note: We treat the $n$ dimension as the feature rather than $c$ dimension.}. 
(iii) At the same time, the feature extractor $\mathcal{G}$ extracts a feature map $f^G=\mathcal{G}(I)$ from the query image, where $f^G\in \mathbb{R}^{n\times c}$. 
(iv) The pose $\hat{P}$ is iteratively refined by 
minimizing the feature cosine similarity loss $\mathcal{L}_{feature}$ \cite{chen2022dfnet}~ between $f^{rend}$ and $f^G$:

\begin{equation} \label{eq:featloss}
\mathcal{L}_{feature} =  \sum^{c}_{i=1}\left(1 -\frac{\langle f^{rend}_{:,i}\cdot f^{G}_{:,i} \rangle} {\|f^{rend}_{:,i}\|_{2}\cdot\|f^{G}_{:,i}\|_{2}}\right)
\end{equation}
where $f^{rend}_{:,i},f^{G}_{:,i}\in \mathbb{R}^{n}$, $\langle\cdot,\cdot\rangle$ denotes the inner product between two vectors and $\|\cdot\|_{2}$ represents the L2 norm. Different from the common feature matching literature's \cite{sun2021loftr,Li20} convention, our features are normalized along the spatial direction instead of the channel direction to ensure the consistency of the neighboring pixels. 



Our method can be regarded as post-processing to the initial pose $\hat{P}$. We do not save the updated weights of the APR method since we restart from the initial state when given a new query image.
\input{fig_tex/figure2_nfs_arch}
\subsection{Neural Feature Synthesizer} \label{sec:NFS}
We propose a Neural Feature Synthesizer (NeFeS) model that directly renders dense feature maps of a given viewpoint to refine the predictions of an underlying APR network. Similar to NeRF-W \cite{martinbrualla2020nerfw}, our NeFeS architecture uses a base MLP module with \textit{static} and \textit{transient} heads that predict the static and transient density ($\sigma^{(s)}$ and $\sigma^{(\tau)}$) and view-dependent color ($c^{(s)}$ and $c^{(\tau)}$) respectively, given an input 3D position ($\mathbf{x}$) and viewing direction ($\mathbf{d}$). We use the frequency encoding \cite{vaswani2017attention,Mildenhall20} to encode all 3D positions and view directions. The transient head models the colors of the 3D points using an isotropic normal distribution and predicts a view-dependent variance value ($\beta^2$) for the transient color distribution.
To render the color of a given pixel, the original volume rendering formulation in NeRF \cite{Mildenhall20} is augmented to include the transient colors and densities, and the color of a given image-pixel ($\hat{\mathbf{C}}(\mathbf{r})$) is computed as a composite of the static and transient components. Here, $\mathbf{r}$ denotes the ray (corresponding to the pixel) on which points are sampled to compute the volume rendering quadrature approximation \cite{Max95}. The variances of sampled points along the corresponding ray are also rendered using only the transient densities (and \textit{not} the static densities) to obtain a per-pixel color variance $\beta(\mathbf{r})^2$. We refer reader to \cite{martinbrualla2020nerfw} for more mathematical details on the static+transient volume rendering formulation. 

We expand the output of the static MLP to also predict features for an input 3D position. The output dimension is $N_c + N_f$, where $N_f$ features are predicted along with RGB values. The per-pixel features are rendered using the same volume rendering quadrature approximation \cite{Max95}:
\begin{equation}
    \begin{aligned} \hat{\mathbf{F}}_f(\mathbf{r}) = \sum_{i=1}^N T_i\left(1-\exp \left(-\sigma^{(s)}_i \delta_i\right)\right) \mathbf{f}_i, \\ T_i=\exp \left(-\sum_{j=1}^{i-1} \sigma^{(s)}_j \delta_j\right)\end{aligned}
\end{equation}
where $\mathbf{f}_i$ and $\sigma^{(s)}_i$ are the feature and density predicted by the static MLP for a sampled point on the ray, and $\delta_i$ is the distance between sampled quadrature points $i$ and $i+1$.

\cref{fig:NFS_arch} demonstrates the architecture of our proposed NeFeS model. We propose two crucial components in the rendering pipeline of our NeFeS architecture that ensure the robustness of our rendered features.

\textbf{Exposure-adaptive ACT.} In the context of camera relocalization, testing images may differ in exposure or lighting from training sequences. To address this, DFNet \cite{chen2022dfnet} proposed using the luminance histogram of the query image as a latent code input to the color prediction head of the NeRF MLP. However, since our NeFeS outputs both colors and features simultaneously, we find this approach perturbs the feature output values and causes instability. Ideally, the feature descriptors should be able to maintain local invariance even under varying exposure. Inspired by Urban Radiance Fields (URF) \cite{rematas22urban}, we propose to use an \textit{exposure-adaptive Affine Color Transformation (ACT)} which is a $3\times3$ matrix $\mathbf{K}$ and a $3$-dimensional bias vector $\mathbf{b}$ predicted by a $4$-layer MLP with the query image's luminance histogram $\mathbf{y}_I$. Unlike URF, which uses a pre-determined exposure code, we use the query image's histogram embedding for accurate appearance rendering of unseen testing images.
The final per-pixel color $\hat{\mathbf{C}}(\mathbf{r})$ is computed using the affine transformation as $\hat{\mathbf{C}}(\mathbf{r}) = \mathbf{K}\hat{\mathbf{C}}_{rend}(\mathbf{r}) + \mathbf{b}$, where $\hat{\mathbf{C}}_{rend}(\mathbf{r})$ is the rendered per-pixel color obtained using the static and transient MLPs.

\textbf{Feature Fusion Module} We propose a Feature Fusion module to fuse the rendered colors and features to produce the final feature map. The rendered colors and features are concatenated and fed into the fusion module consisting of three 3x3 convolutions, followed by a 5x5 convolution and a batch normalization layer. During inference, we render colors and features for all $H\!\times\!W$ image pixels and the resulting $H\!\times\!W\!\times\!(N_c+N_f)$ tensor is processed by the module. Note, for efficiency during training, we sample $S\!\times\!S$ regions to render and apply the loss only to those pixels each iteration.

We use $\mathcal{H}$ to represent the fusion module. The final output feature result is:
\begin{equation}
    \hat{\mathbf{F}}_{fusion}(\mathcal{R}) = \mathcal{H}(\hat{\mathbf{C}}(\mathcal{R}), \hat{\mathbf{F}}_f(\mathcal{R}))
\end{equation}
where $\mathcal{R}$ is the sampled region as described above. 

We experimentally find that the fusion module produces more robust features than the input rendered features $\hat{\mathbf{F}}_f$. We refer readers to the supplementary for detailed ablations.

\textbf{Training the Feature Synthesizer} \label{sec:training}
The high-level concept of training the NeFeS is motivated by feature field distillation proposed in \cite{Kobayashi22}, which essentially distills the 2D backbone features into a 3D NeRF model. However, 2D features in our NeFeS need to be closely related to the direct matching formulation \cite{Irani99,chen21}. In this work, we use the trained 2D feature extractor from \cite{chen2022dfnet}~ to produce the feature labels due to its effectiveness in generating domain invariant features. 

\input{table_tex/table1_cambridge_main.tex}

\textbf{Loss Functions.} The total loss used to train our NeFeS model consists of a photometric loss $\mathcal{L}_{rgb}$ and two $l_1$-based feature-metric losses:
\begin{equation}
    \mathcal{L} = \mathcal{L}_{rgb} + \lambda_1\mathcal{L}_{f} +  \lambda_2\mathcal{L}_{fusion}, 
    \label{eq:L_NFS}
\end{equation}
The photometric loss is defined as the negative log-likelihood of a normal distribution with variance $\beta(\mathbf{r})^2$:
\begin{equation} \label{eq:nerfw_color_loss}
    \begin{aligned} \mathcal{L}_{rgb}(\mathbf{r})= & \frac{1}{2 \beta_i(\mathbf{r})^2}\left\|\mathbf{C}(\mathbf{r})-\hat{\mathbf{C}}(\mathbf{r})\right\|_2^2 \\ & +\frac{1}{2} \log \beta(\mathbf{r})^2+\frac{\lambda_s}{K} \sum_{k=1}^K \sigma_k^{(\tau)}\end{aligned}
\end{equation}
where $\mathbf{r}$ is the ray direction corresponding to an image pixel, $\mathbf{C}_i(\mathbf{r})$ and $\hat{\mathbf{C}}_i(\mathbf{r})$ are the ground-truth and rendered pixel colors. The third term in \cref{eq:nerfw_color_loss} is a sum of the transient densities of all the points on ray $\mathbf{r}$ and 
is used to ensure that transient densities are sparse.

The feature losses are simply $l_1$ losses:
\begin{equation} \label{eq:featloss_1}
\mathcal{L}_{f}=\sum_{\mathbf{r} \in \mathcal{R}}\|\hat{\mathbf{F}}_f(\mathbf{r})-\mathbf{F_{img}}(I,\mathbf{r})\|_1.
\end{equation}
and
\begin{equation} \label{eq:featloss_2}
\mathcal{L}_{fusion}=\sum_{\mathbf{r} \in \mathcal{R}}\|\hat{\mathbf{F}}_{fusion}(\mathbf{r})-\mathbf{F_{img}}(I,\mathbf{r})\|_1.
\end{equation}
where $\mathbf{F_{img}}(I,\cdot)$ are the features extracted from the training images using the pre-trained 2D feature extractor \cite{chen2022dfnet}. Note that, $\mathcal{L}_f$ is applied to the rendered features $\hat{\mathbf{F}}_f$ and $\mathcal{L}_{fusion}$ is applied to the fused features $\hat{\mathbf{F}}_{fusion}$. We experimentally find that using $l_1$ gives more robust features than $l_2$ and cosine feature loss for the test time refinement.

\textbf{Progressive Training.} We propose using a progressive schedule to train the NeFeS model. We first train the color and density part of the network for $T_1$ epochs to bootstrap the correct 3D geometry for the network. For these epochs, only $\mathcal{L}_{rgb}$ is used. Then we add $\mathcal{L}_{f}$ with weight $\lambda_1$ for the next $T_2$ epochs to train the feature part of the static MLP. Since the ground-truth features may not be fully multi-view consistent, we apply \textit{stop-gradients} to the predicted density for the feature rendering branch. And finally, we add the feature fusion loss $\mathcal{L}_{fusion}$ with weight $\lambda_2$ for the last $T_3$ epochs. Since the feature fusion module takes both RGB images and 2D features as input, we randomly sample $N_{crop}$ patches of $S\!\times\!S$ regions of the image and features to increase training efficiency. According to our experiments, this progressive training schedule leads to better convergence and performance. In addition, we apply semantic filtering to improve the network training results. Specifically, we use an off-the-shelf panoptic segmentation method \cite{cheng22mask2former} to mask out temporal objects in the scene such as people and moving vehicles.

\subsection{Direct Pose Refinement} \label{sec:PP_Pose}
While our method is primarily designed to optimize APR, it is also possible to directly optimize camera pose parameters. We explore this feature by showing a possible scenario wherein the source of the pose estimation is either a black box or cannot be optimized (e.g.\ the initial camera pose comes from image retrieval). In these settings, we can set up our proposed method to directly refine the camera poses. Specifically, given an estimated camera pose $\hat{P}= [\mathbf{R}|\mathbf{t}]$, where $\mathbf{R}$ is rotation and $\mathbf{t}$ is the translation component, our method optimizes the camera poses using tangent space backpropagation\footnote{We use the LieTorch \cite{Teed21lietorch} library for this.}. Additionally, we found that using two different learning rates for the translation and rotation parts helps achieve faster and more stable convergence for camera pose refinement. This is different from the standard convention used in \cite{wang2021nerfmm,lin21barf,Chng22garf,bian22nope}~. We refer our readers to supplementary material for more details.

%% file: fig_tex/figure2_nfs_arch.tex
\begin{figure*}[ht]
    \centering
    \includegraphics[width=.8\linewidth]{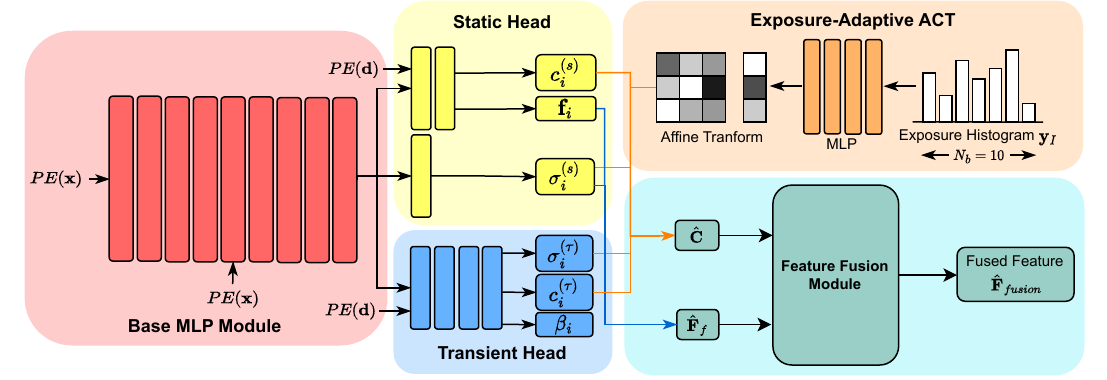}
    \caption{The architecture of our proposed NeFeS model. The query 3D position $\mathbf{x}$ is fed to the network after positional encoding $PE(\cdot)$. The network then splits into two heads: the static head and the transient head. Given a viewing direction $\mathbf{d}$, the rendered color map is generated by fusing static RGB value $c^{(s)}_{i}$, the transient RGB value $c^{\tau}_{i}$ and their corresponding density values $\sigma^{(s)}_{i}$ and $\sigma^{\tau}_{i}$, while the rendered feature map is formed only by static features $\mathbf{f}_{i}$ and density $\sigma^{(s)}_{i}$. In addition, the color map adopts exposure-adaptive ACT to compensate for exposure differences between images. The final feature map $\hat{\mathbf{F}}_{fusion}$ is the concatenation of rendered RGB and feature map processed by the feature fusion module.}
    \label{fig:NFS_arch}
    \vspace{-10pt}
\end{figure*}

%% file: table_tex/table1_cambridge_main.tex
\begin{table*}[!t]
\centering
\small
\begin{tabular}{l|cccc|cc}
\toprule
Methods            & Kings     & Hospital  & Shop      & Church    & Average \\
\midrule
PoseNet(PN)\cite{Kendall15}       & 1.66/4.86 & 2.62/4.90 & 1.41/7.18 & 2.45/7.96 & 2.04/6.23\\
PN Learn $\sigma^2$\cite{Kendall17} & 0.99/1.06 & 2.17/2.94 & 1.05/3.97 & 1.49/3.43 & 1.43/2.85\\
geo. PN\cite{Kendall17}            & 0.88/1.04 & 3.20/3.29 & 0.88/3.78 & 1.57/3.32 & 1.63/2.86\\
LSTM PN\cite{Walch17}            & 0.99/3.65 & 1.51/4.29 & 1.18/7.44 & 1.52/6.68 & 1.30/5.51\\
MapNet\cite{Brahmbhatt18}             & 1.07/1.89 & 1.94/3.91 & 1.49/4.22 & 2.00/4.53 & 1.63/3.64\\
TransPoseNet\cite{Shavit21}       & 0.60/2.43 & 1.45/3.08 & 0.55/3.49 & 1.09/4.94 & 0.91/3.50\\
MS-Transformer\cite{Shavit21multiscene} & 0.83/1.47 & 1.81/2.39 & 0.86/3.07 & 1.62/3.99 & 1.28/2.73\\
MS-Transformer+PAE\cite{Shavit22PAE} & - & - & - & - & 0.96/2.73\\
DFNet \cite{chen2022dfnet}      & 0.73/2.37 & 2.00/2.98 & 0.67/2.21 & 1.37/4.03 & 1.19/2.90\\
DFNet + $\textbf{NeFeS}_{\textbf{50}}$\textbf{(ours)}    & \textbf{0.37}/\textbf{0.54} & \textbf{0.52}/\textbf{0.88} & \textbf{0.15}/\textbf{0.53} & \textbf{0.37}/\textbf{1.14} & \textbf{0.35}/\textbf{0.77}\\
\bottomrule
\end{tabular}
\caption{\textbf{Comparisons on Cambridge Landmarks.} We compare our proposed test-time refinement method with single-frame APR methods. The subscript of $\textbf{NeFeS}_{\textbf{50}}$ denotes the number of optimization iterations used for APR refinement. We report the median position and orientation errors in $m/\degree$. The best results are highlighted in \textbf{bold}.}
\label{table:1}
\vspace{-10pt}
\end{table*}

%% file: sec/04_exp.tex
\section{Experiments} \label{sec:experiments}
We implement our method in PyTorch \cite{paszke2019pytorch}. 
Implementation details about the NeFeS architecture, progressive training scheduling, and pose refinement can be found in the supplementary\footnote{Supplementary: Implementation Details}.

\input{table_tex/table2_7scenes_main.tex}
\subsection{Evaluation on Cambridge Landmarks}
We evaluate our proposed refinement method on Cambridge Landmarks \cite{Kendall15}~, which is a popular outdoor dataset used for benchmarking pose regression methods. The dataset contains handheld smartphone images of scenes with large exposure variations and covers an area of 875$m^2$ to 5600$m^2$. The training sequences contain 200-1500 samples, and test sets are captured from different sequences. 
For each test image, we refine the model using the approach in \cref{sec:PP_APR} for $m=50$ iterations.

We first test our method on top of an open-sourced SOTA single-frame APR method. \cref{table:1} summarizes the results of our method and existing APR methods. Our method achieves the best accuracy across all four scenes when coupled with the DFNet. In \cref{sec:generalization}, we demonstrate the performance of our method with other APR approaches. Particularly, our method improves DFNet by as much as 70.6\% compared to its scene average results. All the per-scene performances from the other compared methods are taken from their papers, except for MS-Transformer+PAE, which only reports the scene average median errors. We encourage our readers to check out supplementary for more thorough comparisons and ablations to the Cambridge Landmarks dataset.
\input{fig_tex/figure3_7scenes_dslam_vs_colmap}

\subsection{Evaluation on 7-Scenes}

We further evaluate our method on Microsoft 7-Scenes dataset \cite{Glocker13,Shotton13}, which includes seven indoor scenes ranging in size from $1m^3$ to $18m^3$ and up to 7000 training images for each scene. We sub-sampled the large training sequences in the dataset to 1000 images for training our NeFeS model. 
The original `ground truth' (GT) poses are obtained from RGB-D SLAM (dSLAM) \cite{Newcombe11KinectFusion}. However, we observe imperfection in the GT poses due to the asynchronous data between the RGB and depth sequences, and this results in low-quality NeRF renderings, as shown in \cref{fig:7scenes_dslam_vs_colmap}. Thus, we use alternative `ground truth' provided by \cite{brachmann2021limits} for our experiments. The authors \cite{brachmann2021limits} demonstrate that their camera poses reconstructed by COLMAP \cite{schoenberger2016sfm}, a structure-from-Motion (SfM) library, are more accurate for image-based relocalization. We refer the reader to our supplementary \footnote{Supplementary: 7-Scenes Dataset Details} for more details about the GT poses.

For a fair comparison, we use the SfM poses to re-train baseline APR methods using their official code, except for PoseNet, in which we use the open-sourced code from \cite{chen21}. We trained each APR method 3-4 times to select the best-performing model. 
We use DFNet + NeFeS$_{50}$ with the same settings as in Cambridge for our pose refinement experiment. We achieve state-of-the-art results (59\%+ better in scene average) in all scenes by running 50 optimization steps (see \cref{table:2}). The results by using the dSLAM ground-truth poses are included in the supplementary\footnote{Supplementary: APR Comparisons with dSLAM GT}, where we also achieve SOTA accuracy.

It is imperative to note that our method is not constrained by the number of optimization steps for the refinement process. In our experience, nearly 50\% of the entire pose error improvement is accomplished within the initial 10 steps. It is up to the user to find the optimal trade-off that fits their computational budget. A detailed analysis of this facet is provided in \cref{sec:iteration_ablation}.

\subsection{Refinement for Different APRs} \label{sec:generalization}
\input{table_tex/table3.tex}
\input{table_tex/table5.tex}

\input{fig_tex/figure6_disturb_test} 
\cref{table:3} shows the results of our method with different APR architectures. Our proposed method exhibits versatility, operating beneficially under various APR architectures, such as PoseNet (classic pose regression architecture), MS-Transformer (EfficentNet CNN backbones with transformer blocks), and DFNet (multi-task network that predicts domain invariant features and poses). 
A full table with per-scene results is provided in Supplementary. 


\subsection{Optimize APR vs. Optimize Pose}

Besides boosting APR, our proposed approach can also refine the initial coarse camera pose, as outlined in \cref{sec:PP_Pose}. We first show a use case of this scenario by coupling our method with image retrieval, where the initial pose can only be optimized due to the non-differentiable nature of the retrieval process. Given a query image, we retrieve its nearest neighbor from the training data using NetVLAD \cite{arandjelovic2016netvlad} and use the associated pose as the initial pose. We set the learning rate to be $lr_R$ and $lr_t$ for rotation and translation components, respectively. Specifically, for indoor scenes, we set $lr_R=0.0087$ (corresponds to $0.5\degree$ in radiance) and $lr_t=0.01$. For outdoor scenes, we set $lr_R=0.01$ and $lr_t=0.1$. 
\cref{table:5} summarises the experimental results, indicating substantial improvements in pose accuracy over the NetVLAD retrieved coarse pose, exceeding the performance of many prior APR approaches.

We further conducted a controlled experiment to investigate whether performance disparities exist between two types of optimization: optimizing the APR's parameters or directly optimizing the pose itself. We evaluated both modes on the DFNet with NeFeS$_{50}$ refinement, as illustrated in \cref{table:4}. The result suggests that while both refinement approaches can effectively improve the pose accuracy, optimizing the neural network's parameters obtain better result than directly optimizing the pose itself, which is an interesting insight. Nevertheless, optimizing the pose remains is also valuable, particularly when the initial pose is derived from a non-differentiable or a black-box pose estimator.

\input{table_tex/table4.tex}

\subsection{Pose Refinement Bounds}

Our proposed refinement method relies on matching rendered features with query image features during test time,
so it may fail when there is not sufficient overlap between the two feature maps. 
To determine the bounds of our refinement method, we randomly perturb the ground-truth pose and determine the maximum perturbation at which our method stops converging. We jitter the orientation or position of the ground truth pose components separately while gradually increasing the magnitude of the perturbation. We use two scenes, an indoor scene (\textit{7-Scenes: Heads}) and an outdoor scene (\textit{Cambridge: Shop Facade}), to illustrate our results in \cref{fig:perturb_test}. We observe that our method cannot refine pose errors larger than $35\degree$. In case of translational errors, our method can refine errors up to $0.6m$ on \textit{Heads} (indoor scene) and up to $4m$ in \textit{Shop Facade} (outdoor scene). This difference may come from the discrepancy in scale between indoor and outdoor settings. For example, in the small-scale scene of \textit{Heads}, the camera is closer to the objects, hence even small movements lead to a large change in the rendering.

\subsection{NeFeS Ablation} \label{sec: nfs_ablation}
This section presents the ablation study of our NeFeS network. In \cref{table:6}a, we gradually remove the exposure-adaptive ACT and the Feature Fusion module and evaluate their impact on the performance of our approach on \textit{Cambridge Shop Facade}. The results demonstrate that the removal of each component leads to a deterioration in pose accuracy, indicating the effectiveness of both components. A noteworthy insight from our architecture design is that the superior pose estimation accuracy is attributed to integrating both Exposure-adaptive ACT and the Feature Fusion module. The feature fusion module can smooth potential noises (outliers) from directly-rendered 3D features.

In \cref{table:6}b, we compare our progressive training scheduling with the combined scheduling, where all three loss terms have been enabled simultaneously since the beginning of the training. The results reveal that the progressive training scheduling results in better accuracy, providing further support for our design decisions.

\input{table_tex/table6}

\subsection{Number of Iterations vs. Accuracy Trade-off} \label{sec:iteration_ablation}
\input{fig_tex/figure5_iter_test}
In \cref{fig:iter_test}, we plot the relationship between the number of optimization iterations and pose error. Both translation and rotation errors reduce significantly in only $10$ iterations. The errors start to plateau around $50$ steps. Although we can achieve even lower errors with more iterations, we think this strikes a balance between accuracy and efficiency, and explains how we set our previous experiments.

\subsection{Spatial vs Channel-wise Normalization}
As described in \cref{sec:PP_APR}, we empirically find spatial-wise normalized features yield higher accuracy than channel-wise normalized features when computing the feature cosine similarity loss $\mathcal{L}_{feature}$, evidenced by \cref{supp:table:channel_spatial_norm}. This is likely due to our spatially sensitive dense direct matching, akin to methods like Direct Sparse Odometry \cite{Engel17}. In contrast, channel-wise normalization can introduce inconsistencies among neighboring pixels.
\input{supp/table_tex/table6}

%% file: table_tex/table2_7scenes_main.tex
\begin{table*}[t]

\centering
\resizebox{\textwidth}{!}{
\begin{tabular}{l|ccccccccccccc|c}
\toprule
Methods                                       & Chess     && Fire        && Heads       && Ofﬁce       && Pumpkin     && Kitchen     && Stairs      & Average \\

\midrule
PoseNet                      & 0.10/4.02 && 0.27/10.0   && 0.18/13.0   && 0.17/5.97   && 0.19/4.67   && 0.22/5.91   && 0.35/10.5   & 0.21/7.74 \\
MapNet                   & 0.13/4.97 && 0.33/9.97   && 0.19/16.7   && 0.25/9.08   && 0.28/7.83   && 0.32/9.62   && 0.43/11.8   & 0.28/10.0 \\
MS-Transformer       & 0.11/6.38 && 0.23/11.5   && 0.13/13.0   && 0.18/8.14   && 0.17/8.42   && 0.16/8.92   && 0.29/10.3   & 0.18/9.51 \\
PAE                                 & 0.13/6.61 && 0.24/12.0   && 0.14/13.0   && 0.19/8.58   && 0.17/7.28   && 0.18/8.89   && 0.30/10.3   & 0.19/9.52 \\
DFNet                                & 0.03/1.12 && 0.06/2.30   && 0.04/2.29   && 0.06/1.54   && 0.07/1.92   && 0.07/1.74   && 0.12/2.63   & 0.06/1.93 \\

DFNet + $\textbf{NeFeS}_{\textbf{50}}$\textbf{(ours)}   & \textbf{0.02}/\textbf{0.57} && \textbf{0.02}/\textbf{0.74} && \textbf{0.02}/\textbf{1.28} && \textbf{0.02}/\textbf{0.56} && \textbf{0.02}/\textbf{0.55} && \textbf{0.02}/\textbf{0.57}   && \textbf{0.05}/\textbf{1.28} & \textbf{0.02}/\textbf{0.79} \\

\bottomrule
\end{tabular}
}
\caption{\textbf{Comparisons on 7-Scenes dataset.} We compare the proposed refinement method with previous single-frame APR methods. We evaluate all methods with SfM ground truth poses provided in \cite{brachmann2021limits}, measured in median translational error (m) and rotational error (\degree). Numbers in \textbf{bold} represent the best performance.}
\label{table:2}
\end{table*}

%% file: fig_tex/figure3_7scenes_dslam_vs_colmap.tex
\begin{figure}[t]
    \centering
   \begin{subfigure}{0.33\linewidth}
        \centering
       \includegraphics[width=\linewidth]{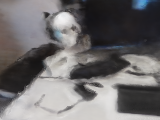}
   \end{subfigure}%
   \begin{subfigure}{0.33\linewidth}
       \centering
       \includegraphics[width=\linewidth]{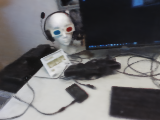}
   \end{subfigure}\vspace{-1mm}
   \begin{subfigure}{0.33\linewidth}
       \centering
       \includegraphics[width=\linewidth]{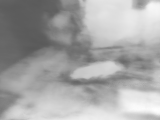}
       \caption{dSLAM NeRF}
   \end{subfigure}%
   \begin{subfigure}{0.33\linewidth}
       \centering
       \includegraphics[width=\linewidth]{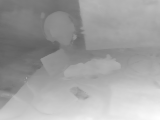}
       \caption{SfM NeRF}
   \end{subfigure}
\caption{Qualitative comparison between the NeRFs trained by dSLAM GT pose (a) vs. SfM GT pose (b). As illustrated, SfM NeRF (PSNR 19.94 dB) can render superior geometric details (bottom row) than dSLAM NeRF (PSNR 16.11 dB).}
\label{fig:7scenes_dslam_vs_colmap}
\end{figure}

%% file: table_tex/table3.tex
\begin{table}[t]
\centering
\small
\begin{tabular}{l|cc}
\toprule
Dataset               & 7-Scenes      &  Cambridge\\
\midrule
PoseNet               & 0.21m/7.74     & 2.04m/6.23\\
\makecell[c]{+ Ours}  & \textbf{0.08m}/\textbf{2.83\degree} & \textbf{0.54m}/\textbf{1.05\degree} \\
\midrule
MS-Trans.             & 0.18m/9.51\degree     & 1.28m/2.73\degree\\
\makecell[c]{+ Ours}  & \textbf{0.11m}/\textbf{3.46\degree} & \textbf{0.43m}/\textbf{1.04\degree}\\
\midrule
DFNet                 & 0.06m/1.93\degree     & 1.19m/2.90\degree\\
\makecell[c]{+ Ours} & \textbf{0.02m}/\textbf{0.79\degree} & \textbf{0.35m}/\textbf{0.77\degree}\\
\bottomrule
\end{tabular}
\caption{\textbf{Pose refinement on different APR architectures.} Our refinement method can effectively improve pose estimation results for different APR architectures.}
\label{table:3}
\end{table}


%% file: table_tex/table5.tex
\begin{table}[t]
\centering
\small
\begin{tabular}{l|cc}
\toprule
Dataset   & NetVlad     & Optimize Pose \\
\midrule
7-Scenes  & 0.32m/13.72\degree & \textbf{0.14m}/\textbf{3.97\degree}\\
Cambridge & 3.18m/7.74\degree  & \textbf{1.15m}/\textbf{1.30\degree}\\
\bottomrule
\end{tabular}
\caption{\textbf{Pose refinement on NetVlad} Our method also works on poses initialized with non-APR-based methods, such as NetVlad image retrieval. Since the initial pose error is relatively large, we refine the poses with 100 iterations.}
\label{table:5}
\end{table}

%% file: fig_tex/figure6_disturb_test.tex

\begin{figure*}[ht]
    \centering
   \begin{subfigure}{.24\linewidth}
        \centering
       \includegraphics[width=\linewidth]{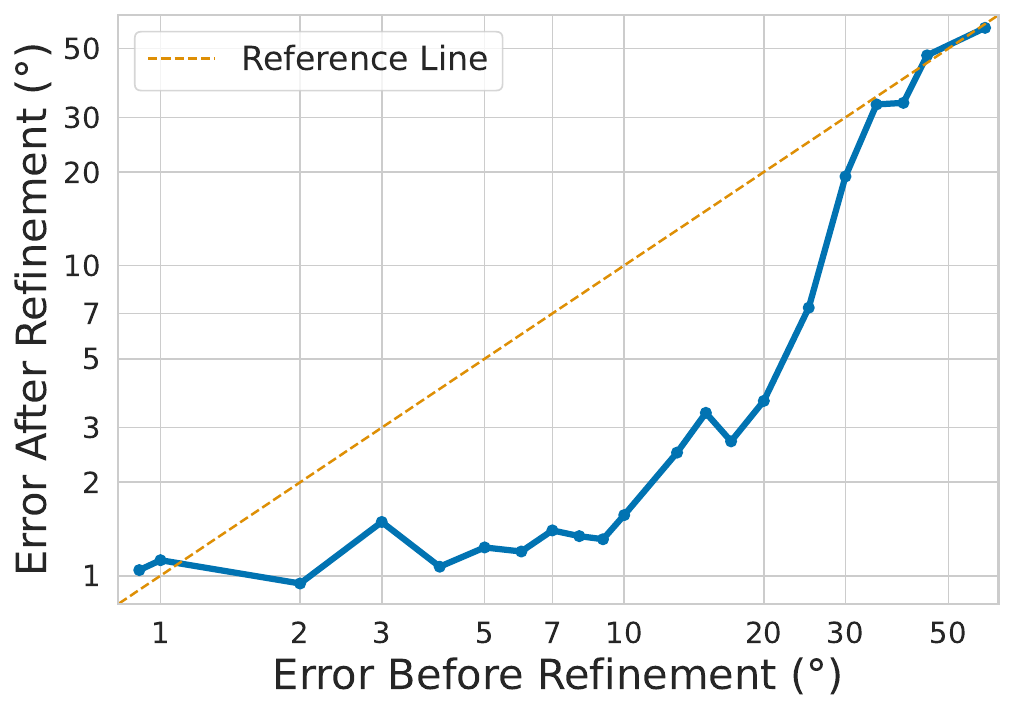}
       \caption{\small{Rotation test (Indoor)}}
   \end{subfigure}\hfill
   \begin{subfigure}{.24\linewidth}
       \centering
       \includegraphics[width=\linewidth]{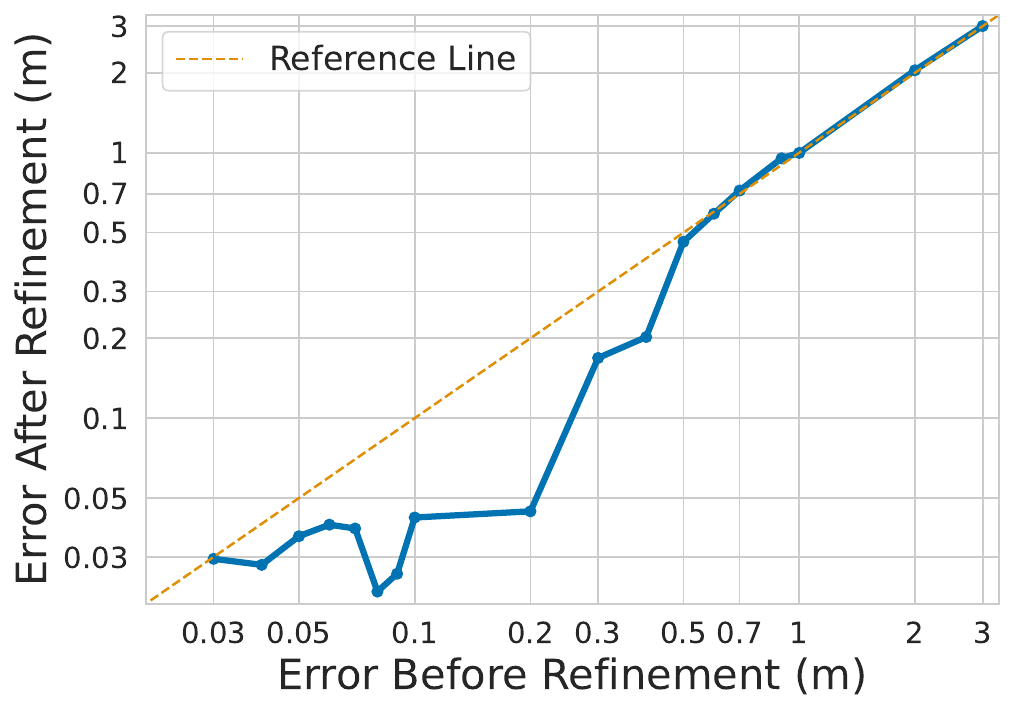}
       \caption{\small{Translation test (Indoor)}}
   \end{subfigure}\hfill
     \begin{subfigure}{.24\linewidth}
        \centering
       \includegraphics[width=\linewidth]{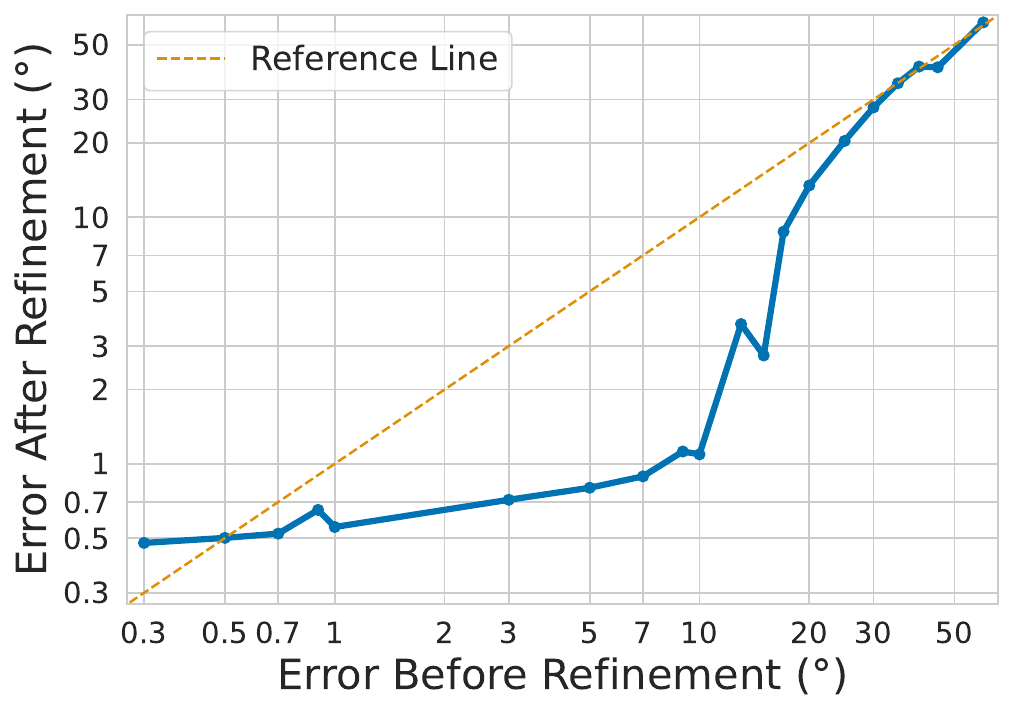}
       \caption{\small{Rotation test (Outdoor)}}
   \end{subfigure}\hfill
   \begin{subfigure}{.24\linewidth}
       \centering
       \includegraphics[width=\linewidth]{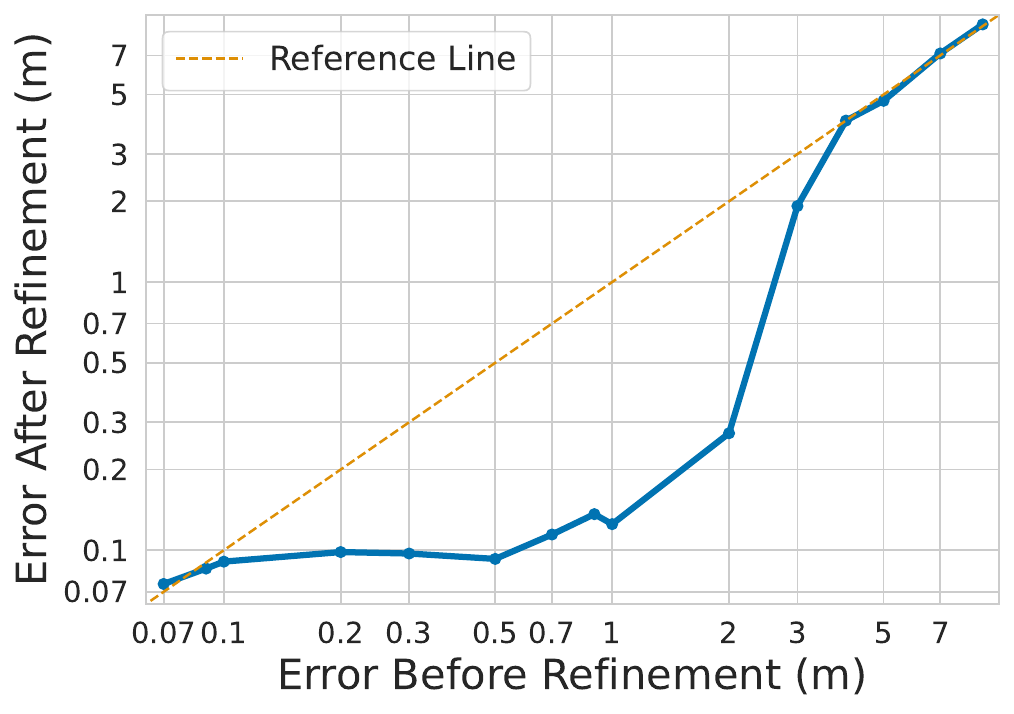}
       \caption{\small{Translation test (Outdoor)}}
   \end{subfigure}
\caption{
Experiments on pose refinement bounds of our method in indoor and outdoor scenes. Each plot shows errors before (x-axis) and after (y-axis) refinement when ground-truth pose is perturbed by varying magnitudes.\ Dashed green line is `$y\!\!=\!\!x$'. Points below this line indicate a reduction in pose error using our refinement method.
}
\label{fig:perturb_test}
\end{figure*}

%% file: table_tex/table4.tex
\begin{table}[t]
\centering
\small
\begin{tabular}{l|ccc}
\toprule
Dataset   & DFNet     & Optimize Pose & Optimize APR \\
\midrule
7-Scenes   & 0.06m/1.93\degree & 0.04m/1.16\degree     & \textbf{0.02m}/\textbf{0.79\degree}    \\
Cambridge & 1.19m/2.90\degree & 0.66m/1.39\degree     & \textbf{0.35m}/\textbf{0.77\degree}   \\
\bottomrule
\end{tabular}
\caption{\textbf{Pose refinement vs. APR refinement} We study on the optimization over APR vs. the pose. Both methods can effectively optimize pose accuracy. However, optimizing APR can obtain lower errors than optimizing poses given the same number of iterations.}
\label{table:4}
\end{table}

%% file: table_tex/table6.tex
\begin{table}[t]

\centering
\resizebox{\linewidth}{!}{

\begin{tabular}{lc}
        \multicolumn{2}{c}{\textbf{(a) NeFeS Architecture Ablation}}\\
        \\
            \toprule
            Method                               & Shop Facade \\
            \midrule
            Initial Pose Error &  0.67m/2.21\degree \\
            Refine w/ NeFeS (ours) & \textbf{0.15m/0.53\degree}  \\
            - Exposure-adaptive ACT \textcircled{A} & 0.15m/1.20\degree  \\
            - \textcircled{A}+Feature Fusion & 0.37m/1.62\degree\\
            
            \bottomrule
        \end{tabular}

\begin{tabular}{lc}
        \multicolumn{2}{c}{\textbf{(b) Training Scheduling Ablation}}\\
        \\
            \toprule
            Method              & Shop Facade \\
            \midrule
            Combined & 0.17m/0.80\degree  \\
            Progressive & \textbf{0.15m/0.53\degree}\\
            \bottomrule
        \end{tabular}

}
\caption{\textbf{(a)} Ablation on NeFeS architecture. \textbf{(b)} Ablation on the proposed training scheduling}
\label{table:6}
\end{table}

%% file: fig_tex/figure5_iter_test.tex
\begin{figure}[t]
    \centering
   \includegraphics[width=\linewidth]{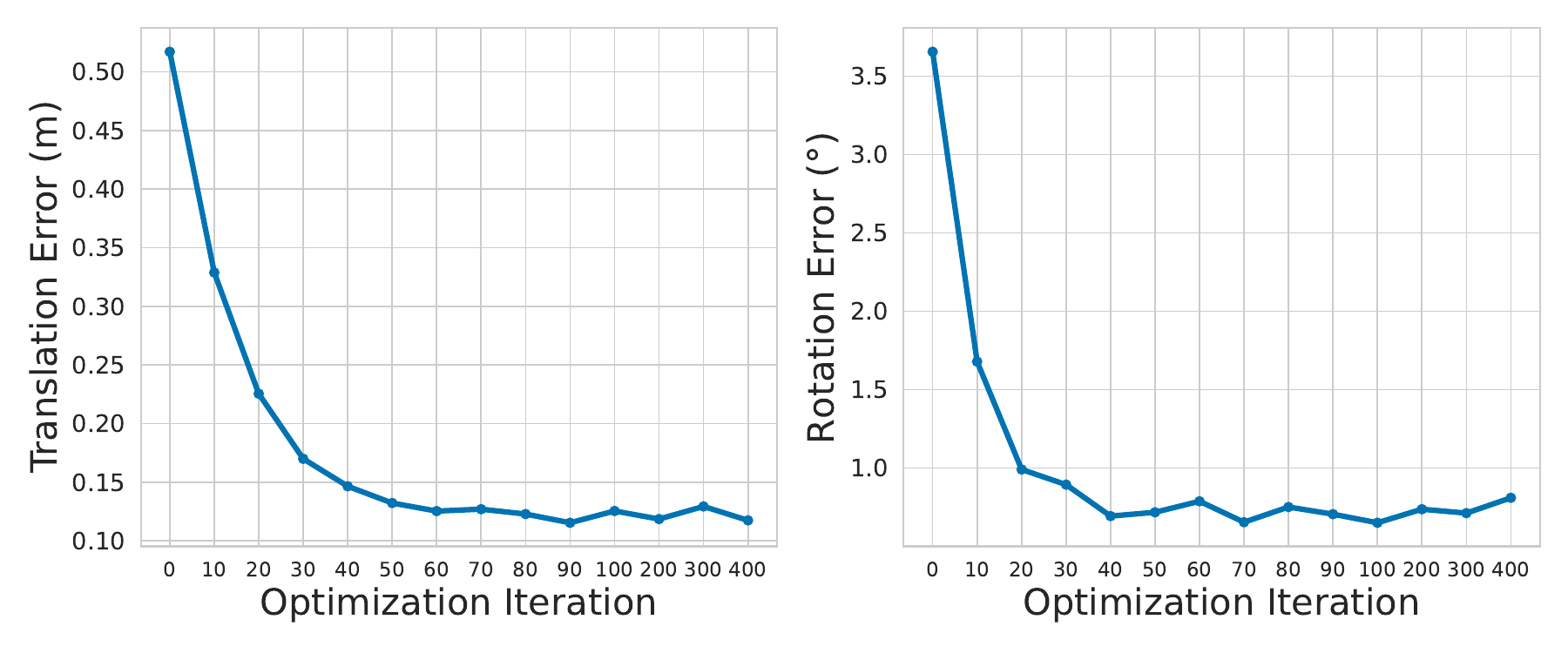}
\caption{Plots of rotation and translation errors against the number of iteration on \textit{Cambridge: Shop Facade} scene.}
\label{fig:iter_test}
\end{figure}

%% file: supp/table_tex/table6.tex
\begin{table}[h]
\centering
\begin{tabular}{lc}
\toprule
Methods & Shop \\
\midrule
Channel                         & 0.20m/1.05$\degree$\\
Spatial                         & \boldred{0.15m}/\boldred{0.53$\degree$} \\
\bottomrule
\end{tabular}
\caption{Performance comparison between using Channel-wise vs. Spatial-wise normalization in loss $\mathcal{L}_{feature}$ during refinement.}
\label{supp:table:channel_spatial_norm}
\end{table}

%% file: sec/05_conclusion.tex
\section{Conclusion} \label{sec:conclusion}

We tackle the camera relocalization problem and improve absolute pose regression (APR) methods by proposing a test-time refinement method. In particular, we design a novel model named Neural Feature Synthesizer (NFS), which can encode 3D geometric features. Given an estimated pose, the NFS renders a dense feature map, compares it with the query image features, and back-propagates the error to refine estimated camera poses from APR methods. In addition, we propose a progressive learning strategy and a feature fusion module to improve the feature robustness of the NFS model. The experiments demonstrate that our method can greatly improve the accuracy of APR methods.
Our method provides a promising direction for improving the accuracy of APR methods.

%% file: supp/supp_00.tex



\section{Supplementary}
\subsection{Implementation Details}

\subsubsection{Architecture details}
The model is trained with the re-aligned and re-centred poses in SE(3), as in \cite{Mildenhall20}. 
We use a coarse-to-fine sampling strategy with 64 sampled points per ray in both stages. 
The width of the MLP layers is $128$ and we output $N_c=3$ and $N_f=128$ in the last layer of the fine stage MLP. 
For the exposure-adaptive ACT module, we compute the query image's histogram $\mathbf{y}_I$ in YUV color space and bin the luminance channel into $N_b=10$ bins. We then feed the binned histogram to 4-layer MLPs with a width of 32. The exposure-adaptive ACT module outputs the exposure compensation matrix $\mathbf{K}$ and the bias $\mathbf{b}$, which directly transform the integrated colors $\hat{\mathbf{C}}_{NFS}(\mathbf{r})$ of the main networks, with negligible computational overhead.
We run the APR refinement process for $m$ iterations per image using the direct feature matching loss $\mathcal{L}_{feature}$ with a learning rate of $1\times10^{-5}$. Our default value for $m$ is 50 unless specified, denoted as NeFeS$_{50}$. The NeFeS model renders features with a shorter side of 60 pixels and then upsample them using bicubic interpolation to 240 for feature matching.

\subsubsection{Progressive training schedule} The training process for the NeFeS network starts with the photometric loss only for $T_1=600$ epochs by setting $\lambda_1 =\lambda_2 = 0$ in Eq.\ (4). The color and density components of the model are trained with a learning rate of $5 \times 10^{-4}$ which is exponentially decayed to $8 \times 10^{-5}$ over 600 epochs. We randomly sample $1536$ rays per image and use a batch size of $4$. After $600$ epochs, we reset the learning rate to $5 \times 10^{-4}$ and switch on the feature loss ($\mathcal{L}_f$ in Eq.\ (6)) for the next $T_2=200$ epochs with $\lambda_1 =0.04, \lambda_2 = 0$. The fusion loss ($\mathcal{L}_{fusion}$ in Eq.\ (7)) is switched on for the last $T_3=400$ epochs with coefficients $\lambda_1 =0.02, \lambda_2 = 0.02$. During the third training stage $T_3$, instead of randomly sampling image rays, we randomly sample $N_{crop}\!=\!7$ image patches of size $S\times S$ where $S\!=\!16$.
To extract image features (i.e.\ $\mathbf{F_{img}}(I, \cdot)$) as pseudo-groundtruth, we use the finest-level features from DFNet's \cite{chen2022dfnet} feature extractor module. We resize the shorter sides of the feature labels to $60$.

\subsection{Refinement for Different APRs Full}
This is the supplementary full table for Section 4.3 of the main paper (\cref{table:Refinemen_for_Diff_APRs_FULL}).
\input{supp/table_tex/table5}
\input{fig_tex/figure4_vis_colmap_gt_pose}
\subsection{Qualitative Comparisons}
In \cref{fig:7scenes_qualitative}, we qualitatively compare the refinement accuracy of different APR methods - namely PoseNet\cite{Kendall15,Kendall16,Kendall17}, MS-Transformer\cite{Shavit21multiscene}, DFNet \cite{chen2022dfnet} - with our method, i.e.\ DFNet+NeFeS$_{50}$. We can observe that our method produces the most accurate poses (compared to ground-truth) and has a significant improvement over DFNet in different scenes such as fire [$1000$-$1500$] and kitchen [$1000$-$1500$].
\input{supp/fig_tex/figure1}

\subsection{7-Scenes Dataset Details}
In Sec.\ 4.2 of the main paper, we mention the difference between the dSLAM-generated ground-truth pose and the SfM-generated ground-truth pose for the 7-Scenes dataset. We provide more details in this section.

\textbf{dSLAM vs. SfM GT pose}
Brachmann \emph{et al.} \cite{brachmann2021limits} identified imperfections in the original `ground-truth' (GT) poses generated by dSLAM in the 7-Scenes dataset. The erroneous GT poses originate from sensor asynchronization between the captured RGB images and depth maps. Therefore, Brachmann et al. employed SfM to regenerate a new set of `ground-truth' poses, which subsequently aligned and scaled to match the dSLAM-derived poses. As described in Sec.\ 4.2 of the main paper, we notice that when trained with the SfM ground-truth poses, the rendering quality of NeRF is noticeably boosted compared with using the dSLAM GT poses. The comparison between the trajectories of two sets of ground-truth poses is visualized in \cref{fig:vis_colmap_gt_pose}. An interesting observation is made based on the results presented in Table 2 of our main paper. We notice DFNet achieves superior performance when trained with SfM-grounded GT data, surpassing its performance as originally reported \cite{chen2022dfnet}. This phenomenon may be attributed to utilizing the improved synthetic dataset generated by NeRF during DFNet's \textit{Random View Synthesis} training.

\textbf{APR Comparisons with dSLAM GT.}
To supplement Table 2 of the main paper, we compare previous methods and our method when trained and evaluated using dSLAM GT poses. The results can be found in \cref{supp:table:7scene_dslam}. Note that the pose error is presented in cm/degree to emphasize the distinctions in translational accuracy. Despite NeFeS models being trained using suboptimal dSLAM GT poses in this experiment which reduces the quality of the feature rendering, our model is able to achieve SOTA performance on single-frame APR comparisons. Notably, Coordinet+LENS \cite{Moreau21} is the only single-frame APR technique that achieves our method's proximate outcomes (on translational error). However, it's pertinent to note that LENS requires several days to train a high-quality NeRF model per scene. In stark contrast, the NeFeS model requires a much shorter training duration of approximately 5-20 hours, accompanied by an inference speed over 110 times faster and obviated the need for manual parameter tuning, making NeFeS a notably more cost-effective prospect.

Furthermore, we experiment to see if the current dSLAM pose results can be improved if a better quality NeFeS model is used. We performed joint optimization of NeFeS and ground truth camera poses during training using the method introduced in NeRF-- -- \cite{wang2021nerfmm}. The outcomes reveal that while the NeFeS model attains an enhanced training PSNR from 23.33dB to 27.88dB and the median translation error improves by $1 cm$, the rotation error worsens by $0.07\degree$ since jointly optimizing the dSLAM GT training poses also slightly shifts the world coordinate system of the radiance fields. This refined model's performance is denoted as DFNet + NeFeS$^{- -}_{50}$, as indicated in \cref{supp:table:7scene_dslam}.

\input{supp/table_tex/table1}

\subsection{Comparison with Other Camera Localization Approaches}
Although our paper mainly focuses on test-time refinement on single-frame APR methods, it is only one family of approaches in camera relocalization (see our Related Work section). In \cref{supp:table:our_vs_other}, we compare geometry-based methods and sequential-based methods for camera localization, as well as adding several other single-frame APR methods, including some without code available publicly to support a more thorough comparison. The results on 7-Scenes dataset are evaluated using the original SLAM ground-truth pose, except methods marked by ``(COLMAP)'', which indicates the results evaluated using the COLMAP ground-truth pose for 7-Scenes. The methods that marked by ``(COLMAP to build 3D model)'' indicates COLMAP generated 3D models are used in training and evaluation.

We show that when compared with sequential-based APR methods, our method achieves very competitive results on Cambridge Landmark dataset and 7-Scenes dataset. In addition, for the first time, we show that a single-frame APR method can obtain accuracy of the same magnitude as 3D geometry-based approaches.


\input{supp/table_tex/table2}

\subsection{Featuremetric vs. Photometric Refinement}
In this section, we study the differences between feature-metric refinement and photometric refinement. Prior literature such as iNeRF \cite{yen2020inerf}, NeRF$--$ \cite{wang2021nerfmm}, BARF \cite{lin21barf}, GARF \cite{Chng22garf}, and NoPe-NeRF \cite{bian22nope}, have attempted to `invert' a NeRF model with photometric loss for pose optimization. 

However, directly comparing our featuremetric method with these methods would not be appropriate due to the following reasons:
\textbf{Firstly}, these methods \cite{wang2021nerfmm, lin21barf, Chng22garf, bian22nope} optimize both camera and NeRF model parameters simultaneously but are unsuitable for complex scenes with large motion (e.g.\ 360\degree scenes) since each frame's camera pose is initialized from an identity matrix.
\textbf{Secondly}, these methods do not effectively handle exposure variations, resulting in suboptimal rendering quality.
\textbf{Thirdly}, even with a coarse camera pose initialization, photometric-based inversion methods cannot prevent drifting in refined camera poses, leading to misalignment with the ground truth poses of testing sequences.

Therefore, for a fair comparison with photometric methods, we define a photometric refinement model as the baseline model to compare with. Specifically, for the baseline model, the main architecture from the NeFeS model is maintained but without the feature outputs, and only the RGB colors $\hat{\mathbf{C}}(\mathbf{r})$ are used for photometric pose refinement. The performance of two cases with photometric refinement are demonstrated in \cref{supp:table:photometric-refinement}: first is a sparse photometric refinement that randomly samples pixel-rays, similar to iNeRF \cite{yen2020inerf} or BARF\cite{lin21barf}-like methods; and the other uses dense photometric refinement, which renders entire RGB images. The results indicate that our featuremetric refinement is more robust than all the photometric refinement baselines, as it achieves lower pose errors after $50$ iterations of optimization.

\input{supp/table_tex/table3}

\subsection{Benefit of splitting $lr_R$ and $lr_t$}
As described in Sec.\ 3.3 of the main paper, we find using different learning rates for translation and rotation components as beneficial for fast convergence when we directly refine the camera pose parameters. In this section, we use a toy experiment to illustrate how we determine to use this strategy. We select $20$\% of \textit{Cambridge: Shop Facade}'s test images and perform direct pose refinement for $20$ iterations using our NeFeS model. In \cref{supp:table:split-LRs}, we compare our \textit{different} learning rate setting with several cases of \textit{same} learning rate settings. The learning rate $lr_{R}=lr_t=0.003$ is used in \cite{lin21barf,Chng22garf} and $lr_{R}=lr_t=0.001$ is used in \cite{wang2021nerfmm,yen2020inerf}. We show that by utilizing a different learning rate strategy, the pose error converges much faster and is more stable for both camera position and orientation.

\input{supp/table_tex/table4}

\subsection{Runtime Analysis}

\textbf{Runtime cost.}
Due to better implementation flexibility, we used an unoptimized version of NeFeS in this study. The pytorch-based NeFeS currently runs at 6.9 fps per image including its backpropagation, which is 3x faster than DFNet's NeRF-Hist \cite{chen2022dfnet} and 110x faster than LENS's NeRF-W \cite{Moreau21}.
It is crucial to emphasize that further optimization can be pursued to attain commercial-level efficiency. For example, NeFeS can potentially be accelerated up to 66x using the C++/CUDA-based \texttt{tiny-cuda-nn} and \texttt{instant-ngp} \cite{muller2022instant} frameworks.

\textbf{Training cost.}
Our NeFeS can be trained in parallel with the APR method such as DFNet and takes roughly the same time as the underlying APR method (\textit{i.e.} \ 5-20 hrs depending on scene size). However, the NeFeS model only needs to be trained \textbf{once} and the same model can be applied to different APR methods.


%% file: supp/table_tex/table5.tex
\begin{table}[t]
\centering
\resizebox{\linewidth}{!}{
\begin{tabular}{l|cc|cc|cc}
\toprule
Dataset               & PoseNet    &  \makecell[c]{+ Ours}  &   MS-Trans. & \makecell[c]{+ Ours}   & DFNet       & \makecell[c]{+ Ours} \\
\midrule
7-Scenes              &\multicolumn{6}{c}{pose error in m/\degree}             \\
\midrule
Chess                 & 0.10/4.02  &  0.04/1.35      & 0.11/6.38        & 0.06/1.96               & 0.03/1.12   & 0.02/0.57    \\
Fire                  & 0.27/10.0  & 0.03/1.20        & 0.23/11.5        & 0.06/2.55               & 0.06/2.30   & 0.02/0.74    \\
Heads                 & 0.18/13.0  & 0.12/7.91       & 0.13/13.0        & 0.09/6.19               & 0.04/2.29  & 0.02/1.28    \\
Office                & 0.17/5.97  & 0.02/0.72       & 0.18/8.14        & 0.05/1.69                        & 0.06/1.54 & 0.02/0.56    \\
Pumpkin               & 0.19/4.67  & 0.06/1.57       & 0.17/8.42        & 0.07/1.85               & 0.07/1.92  & 0.02/0.55    \\
Kitchen               & 0.22/5.91  & 0.02/0.68       & 0.16/8.92        & 0.08/2.31               & 0.07/1.74  & 0.02/0.57    \\
Stairs                & 0.35/10.5  & 0.27/6.35       & 0.29/10.3        & 0.34/7.64               & 0.12/2.63 & 0.05/1.28    \\
\midrule
Average               & 0.21/7.74  & \textbf{0.08}/\textbf{2.83}       & 0.18/9.51  &      \textbf{0.11}/\textbf{3.46} &  0.06/1.93  & \textbf{0.02}/\textbf{0.79}    \\
\midrule
Cambridge             & \multicolumn{6}{c}{pose error in m/\degree}  \\
\midrule
Kings                 & 1.66/4.86 & 0.38/0.56        & 0.83/1.47       & 0.43/0.59                        & 0.73/2.37 & 0.37/0.54    \\
Hospital              & 2.62/4.90 & 1.15/1.30        & 1.81/2.39       & 0.61/1.06               & 2.00/2.98 & 0.52/0.88    \\
Shop                  & 1.41/7.18 & 0.21/0.81        & 0.86/3.07       & 0.18/0.98               & 0.67/2.21 & 0.15/0.53    \\
Church                & 2.45/7.96 & 0.42/1.52        & 1.62/3.99       & 0.48/1.53               & 1.37/4.03 & 0.37/1.14    \\
\midrule
Average               & 2.04/6.23 & \textbf{0.54}/\textbf{1.05}        & 1.28/2.73       &     \textbf{0.43}/\textbf{1.04} & 1.19/2.90 & \textbf{0.35}/\textbf{0.77}  \\
\bottomrule
\end{tabular}
}
\caption{\textbf{Pose refinement on different APR architectures.} Our refinement method can effectively improve pose estimation results for different APR methods. PoseNet is the classic pose regression architecture. MS-Transformer is denoted as MS-Trans., which combines EfficentNet CNN backbones with transformer blocks. DFNet is a multi-task network that predicts domain invariant features and poses.}
\label{table:Refinemen_for_Diff_APRs_FULL}
\end{table}

%% file: fig_tex/figure4_vis_colmap_gt_pose.tex
\begin{figure}[t]
  \centering
    \begin{tabular}{c}
      \includegraphics[width=0.8\linewidth]{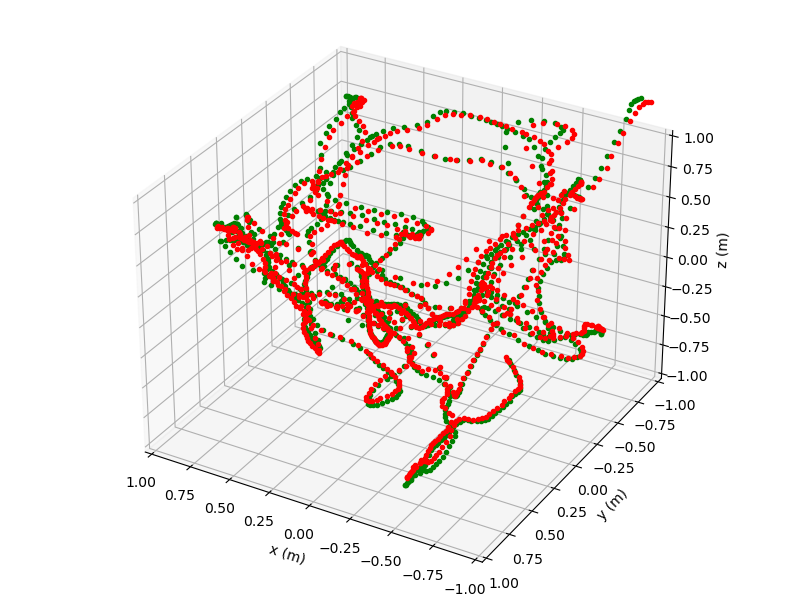}
    \end{tabular}
\caption{A visualization of camera trajectories of \textit{7-Scene: Chess} scene. The original `GT' poses are obtained using dSLAM \cite{Newcombe11KinectFusion} (\textcolor{green}{green}). In this paper, we use SfM GT poses provided by \cite{brachmann2021limits} (\textcolor{red}{red}) for better GT pose accuracy. Two GT trajectories have a median ATE error of 3.5cm/1.46\degree.}

\label{fig:vis_colmap_gt_pose}
\end{figure}

%% file: supp/fig_tex/figure1.tex
\begin{figure*}[h]
  \centering
    \begin{tabular}{cccc}
      \includegraphics[width=0.24\linewidth]{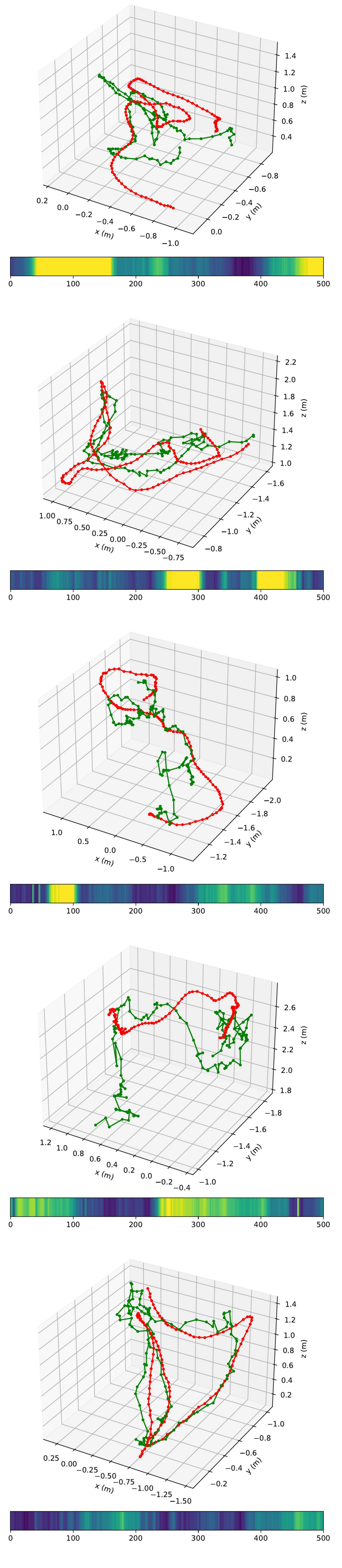} &
      \includegraphics[width=0.24\linewidth]{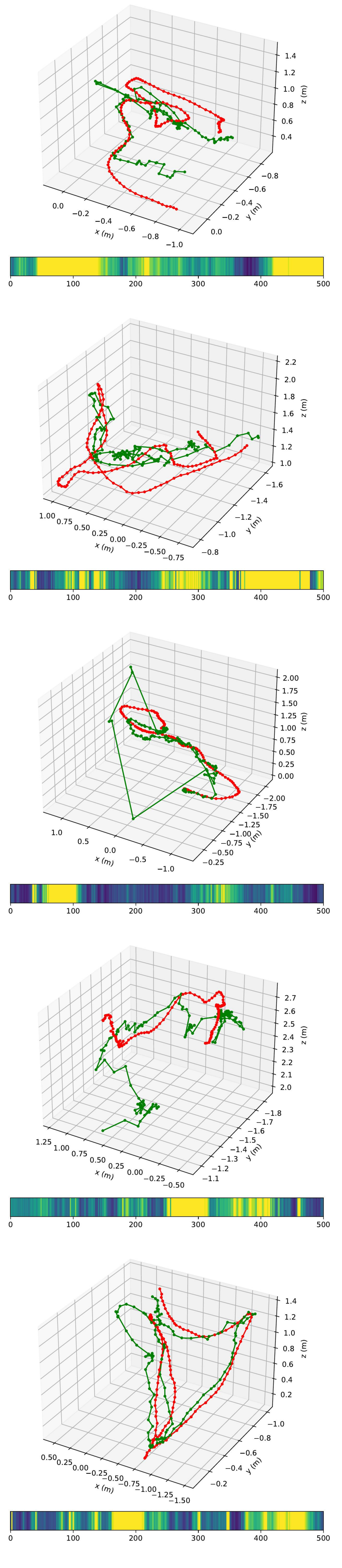}&
      \includegraphics[width=0.24\linewidth]{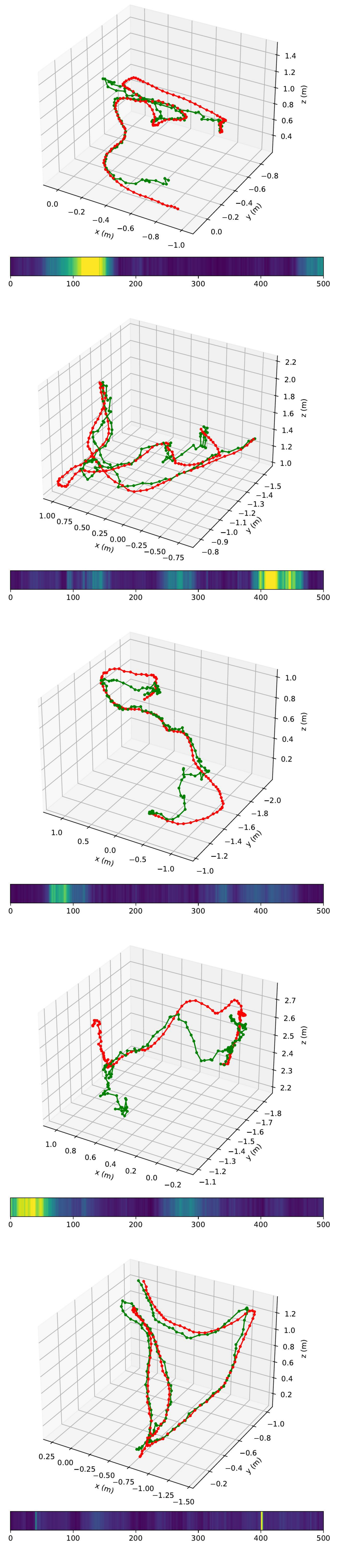}&
      \includegraphics[width=0.24\linewidth]{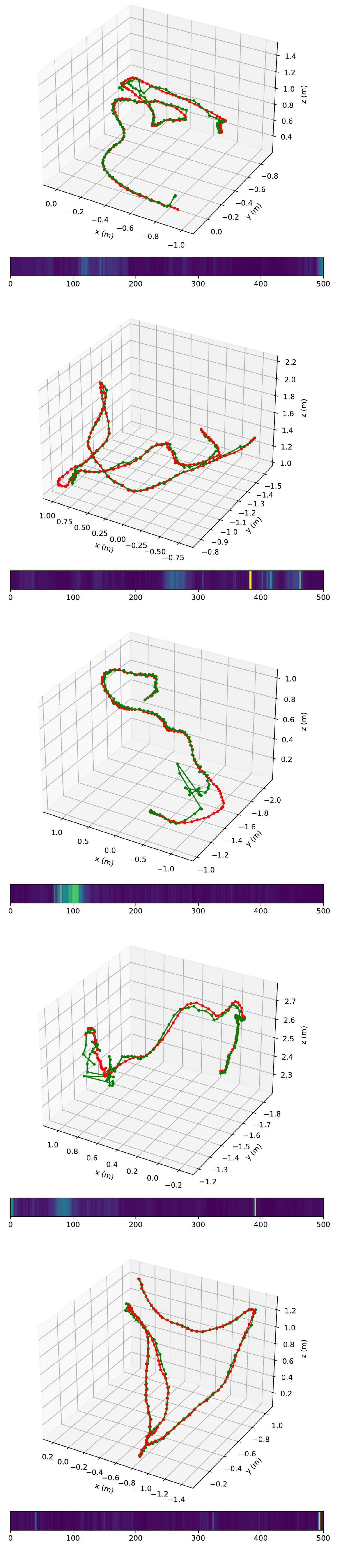}\\
      (a) PoseNet & (b) MS-Transformer & (c) DFNet & (d) \textbf{DFNet+NeFeS$_{50}$}
    \end{tabular}
\caption{Qualitative comparison on the 7-Scenes dataset. The 3D plots show the camera positions: \textcolor{green}{green} for ground truth and \textcolor{red}{red} for predictions. The bottom color bar represents rotational errors for each subplot, where yellow means large error and blue means small error for each test sequence. Sequence names from top to bottom are: fire [$1000$-$1500$], office [$2500$-$3000$], pumpkin [$500$-$1000$], kitchen [$1000$-$1500$], kitchen [$1500$-$2000$].
Since each scene has different numbers of frame, we select $500$ frames from each of them and append the range after scene's name.
}
\label{fig:7scenes_qualitative}
\end{figure*}

%% file: supp/table_tex/table1.tex
\begin{table}[t]
\centering
\resizebox{0.8\linewidth}{!}{
\begin{tabular}{lc}
\toprule
Methods & Average(cm/\degree) \\

\midrule
PoseNet(PN)\cite{Kendall15}         & 44/10.4 \\
PN Learn $\sigma^2$\cite{Kendall17}   & 24/7.87 \\
geo. PN\cite{Kendall17}              & 23/8.12 \\
LSTM PN\cite{Walch17}               & 31/9.85 \\
Hourglass PN\cite{Melekhov17}     & 23/9.53 \\
BranchNet\cite{Wu17}            & 29/8.30 \\
MapNet\cite{Brahmbhatt18}        & 21/7.77 \\
Direct-PN\cite{chen21}           & 20/7.26 \\
TransPoseNet\cite{Shavit21}       & 18/7.78 \\
MS-Transformer\cite{Shavit21multiscene}      & 18/7.28 \\
MS-Transformer+PAE \cite{Shavit22PAE}        & 15/7.28 \\
CoordiNet\cite{moreau2022coordinet}  & 22/9.7\\
CoordiNet+LENS\cite{Moreau21}        & 8/3.0\\
DFNet \cite{chen2022dfnet}          & 12/3.71 \\
\textbf{DFNet + }$\textbf{NeFeS}_{\textbf{50}}$ (dSLAM)  & 8/\textbf{2.80} \\
\textbf{DFNet + }$\textbf{NeFeS}^{- -}_{\textbf{50}}$ (dSLAM)  & \textbf{7}/2.87 \\

\bottomrule
\end{tabular}
}
\caption{We compare the proposed refinement method using 7-Scene dSLAM GT pose \cite{Shotton13} with prior single-frame APR methods, in average of median \textbf{translation error} \textbf{(cm)} and \textbf{rotation error} \textbf{(\degree)}. Numbers in \boldred{bold} represent the best performance.}
\label{supp:table:7scene_dslam}
\end{table}




%% file: supp/table_tex/table2.tex
\begin{table}[t]
\centering
\resizebox{\linewidth}{!}{
\begin{tabular}{c|l|cc}
\toprule
Family &Method & Cambridge & 7-Scenes \\
\midrule
\multirow{1}{*}{\makecell[c]{Seq. 3D}}
& KFNet\cite{Zhou20}       & 13/0.3 & 3/0.88\\
\midrule
\multirow{7}{*}{\makecell[c]{3D}}
& AS\cite{Sattler12} & 29/0.6 & 5/2.5\\
& AS\cite{Sattler17} & 11/0.3 & 4/1.2\\
& DSAC\cite{Brachmann17} & 32/0.8 & 20/6.3\\
& DSAC*\cite{brachmann2020dsacstar} & 15/0.4 & 3/1.4\\
& DSAC*\cite{brachmann2021limits} (COLMAP) & - & 1/0.34\\
& PixLoc\cite{sarlin21pixloc} (COLMAP to build 3D model) & 11/0.3 & 3/1.1\\
& HLoc \cite{sarlin2019HFNet} (COLMAP to build 3D model) & 10/0.2 & 3/1.09\\ 
\midrule
\multirow{4}{*}{\makecell[c]{Seq.\\APR}}
& MapNet+PGO\cite{Brahmbhatt18}        & -          & 18/6.55\\
& AtLoc+\cite{atloc}                   & -         & 19/7.08\\
& TransAPR\cite{Qiao23}                      & 94/2.12  & 17/6.29\\
& VLocNet \cite{Valada18}              & 78/2.82  & 5/3.80\\
\midrule
\multirow{18}{*}{\makecell[c]{1-frame\\APR}}
&PoseNet(PN)\cite{Kendall15}         & 204/6.23 & 44/10.4 \\
&PN Learn $\sigma^2$\cite{Kendall17} & 143/2.85 & 24/7.87 \\
&geo. PN\cite{Kendall17}             & 163/2.86 & 23/8.12 \\
&LSTM PN\cite{Walch17}               & 130/5.51 & 31/9.85 \\
&Hourglass PN\cite{Melekhov17}       & - & 23/9.53 \\
&BranchNet\cite{Wu17}            & - & 29/8.30 \\
&MapNet\cite{Brahmbhatt18}        & 163/3.64 & 21/7.77 \\
&Direct-PN\cite{chen21}           & - & 20/7.26 \\
&TransPoseNet\cite{Shavit21}       & 91/3.50 & 18/7.78 \\
&MS-Transformer\cite{Shavit21multiscene}     & 128/2.73 & 18/7.28 \\
&MS-Transformer+PAE \cite{Shavit22PAE}      & 96/2.73  & 15/7.28 \\
&E-PoseNet \cite{Musallam22}      &  94/2.12 & 17/7/32 \\
& CoordiNet\cite{moreau2022coordinet}  & 92/2.58  & 22/9.7\\
& CoordiNet+LENS\cite{Moreau21}        & 39/1.15  & 8/3.0\\
&DFNet \cite{chen2022dfnet}          & 119/2.90  &12/3.71 \\
&DFNet \cite{chen2022dfnet} (COLMAP)         & -  &6/1.93 \\
&\textbf{DFNet + }$\textbf{NeFeS}_{\textbf{50}}$ & \textbf{35}/\textbf{0.77} & \textbf{8}/\textbf{2.80} \\
&\textbf{DFNet + }$\textbf{NeFeS}_{\textbf{50}}$ (COLMAP) & - & \textbf{2}/\textbf{0.79} \\
\bottomrule
\end{tabular}
}
\caption{This table compares different types of camera relocalization on Cambridge Landmarks and 7-Scenes dataset. We show representative methods for each school of approach: geometry-based methods (3D), sequential-based APR methods (Seq. APR), and single-frame APR methods (1-frame APR). We report the average of median \textbf{translation error} \textbf{(cm)} and \textbf{rotation error ($\degree$)}. Numbers in \textbf{bold} represent the performance of our methods.}
\label{supp:table:our_vs_other}
\end{table}

%% file: supp/table_tex/table3.tex
\begin{table}[t]

\centering
\begin{tabular}{lc}
\toprule
Methods & Hospital \\
\midrule
DFNet                                                   & 2.00m/2.98$\degree$\\
DFNet + Sparse NeRF photometric$_{50}$                         & 1.19m/1.52$\degree$ \\
DFNet + Dense NeRF photometric$_{50}$                         & 0.80m/1.12$\degree$ \\
\textbf{DFNet + }$\textbf{NeFeS}_{\textbf{50}}$         & \boldred{0.52m}/\boldred{0.88$\degree$} \\
\bottomrule
\end{tabular}
\caption{We compare our featuremetric refinement method using the proposed \textbf{NeFeS} network with photometric-based refinement baselines on \textit{Cambridge Hospital}.}
\label{supp:table:photometric-refinement}
\end{table}


%% file: supp/table_tex/table4.tex
\begin{table}[t]
\centering
\small
\begin{tabular}{c|lc}
\toprule
&LR Settings & Shop-20\% (+NeFeS$_{20}$) \\
\midrule
&Initial Pose Error          & 0.58m/3.14$\degree$\\
\multirow{5}{*}{\makecell[c]{Same lr}}&$lr_{R}=lr_t=0.1$              & 0.91m/22.70$\degree$\\
&$lr_{R}=lr_t=0.01$             & 0.49m/1.51$\degree$\\
&$lr_{R}=lr_t=0.003$            & 0.54m/2.44$\degree$\\
&$lr_{R}=lr_t=0.001$         & 0.57m/2.48$\degree$\\
\midrule
\multirow{1}{*}{\makecell[c]{Different lr}}&$lr_R=0.01$, $lr_t=0.1$     & \boldred{0.27m}/\boldred{1.77$\degree$}\\
\bottomrule
\end{tabular}
\caption{We use a toy example to show the benefit of using \textit{different} learning rates over \textit{same} learning rates for translation and rotation components during direct pose refinement. We show four cases for \textit{same} learning rate including two settings that are used in prior works. Our pose refinement results are evaluated by using $20$\% test data of \textit{Cambridge: Shop Facade} and $20$ iterations of optimization.}
\label{supp:table:split-LRs}
\end{table}